%% file: main.tex
\pdfoutput=1

\documentclass[11pt]{article}

\usepackage[]{acl}

\usepackage{times}
\usepackage{latexsym}
\usepackage{tabularx}

\usepackage[T1]{fontenc}

\usepackage[utf8]{inputenc}

\usepackage{microtype}

\input{commands}

\title{ Can Multilinguality benefit Non-autoregressive Machine Translation? }

\author{Sweta Agrawal$^1$ and Julia Kreutzer$^2$ and Colin Cherry$^2$ \\
  $^1$Department of Computer Science, University of Maryland \\
  $^2$Google Research \\
  \texttt{sweagraw@umd.edu}, \texttt{\{jkreutzer, colincherry\}@google.com} \\}

\begin{document}
\maketitle
\begin{abstract}

Non-autoregressive (\nar) machine translation has recently achieved significant improvements, and now outperforms autoregressive (\ar) models on some benchmarks, providing an efficient alternative to \ar inference. However, while \ar translation is often implemented using multilingual models that benefit from transfer between languages and from improved serving efficiency, multilingual \nar models remain relatively unexplored.
Taking Connectionist Temporal Classification (\ctc) as an example \nar model and Imputer as a semi-\nar model \cite {saharia2020non}, we present a comprehensive empirical study of multilingual \nar. We test its capabilities with respect to  positive transfer between related languages and negative transfer under capacity constraints. As \nar models require distilled training sets, we carefully study the impact of bilingual versus multilingual teachers.
Finally, we fit a scaling law for multilingual \nar, which quantifies its performance relative to the \ar model as model scale increases.
\end{abstract}

\input{Introduction}

\input{Method}


\input{ExperimentalSetup}


\input{Results}

\input{Analysis}


\input{Transfer}


\input{Scaling}
\input{RelatedWork}

\section{Conclusion}
While the community has achieved significant progress in improving the performance of \nar and semi-\nar models, the focus on benchmarking on a few language pairs has impeded our understanding of how these models might generalize to multilingual settings. The need for distilled datasets from \ar models  limits the ceiling on translation quality while adding the expense of training additional models to translate the training set. We have shown that generating distilled datasets from a multilingual model is a good alternative, and have provided insights on how multilingual teachers might differ from bilingual ones in quality and complexity. But as we show in our analysis, generating sequences of the right length and with valid tokens remains a challenge, and one that only grows as we move to either multilingual modeling or to more distant language pairs. We are far from using off-the-shelf models to enable non-autoregressive generation in resource constrained scenario even with positive transfer from similar languages. Moving forward, we need to improve the coverage of language pairs studied in order to set better expectations from multilingual \nar models, and to better understand why our multilingual models are unable to reach the quality of their teachers, even with increased capacity.

\bibliography{anthology, custom}
\bibliographystyle{acl_natbib}




\end{document}

%% file: commands.tex
\usepackage{xspace}
\usepackage{colortbl}
\usepackage{paralist}
\usepackage{booktabs}
\usepackage{colortbl}
\usepackage{import}
\usepackage{cmll}
\usepackage{MnSymbol}
\usepackage{multirow}
\usepackage{graphicx}
\usepackage{enumitem}
\usepackage{subcaption}
\usepackage{caption}
\usepackage{rotating}
\usepackage[normalem]{ulem}
\usepackage{pifont}
\newcommand{\cmark}{\ding{51}}%
\newcommand{\xmark}{\ding{55}}%
\usepackage{siunitx} 
\usepackage{multirow, makecell}

\definecolor{darkgreen}{rgb}{0.0, 0.42, 0.24}
\definecolor{green}{RGB}{112, 173,71}
\definecolor{blue}{RGB}{68, 114,196}
\definecolor{orange}{RGB}{237, 125,49}
\definecolor{red}{RGB}{202, 54,49}
\definecolor{yellow}{RGB}{222,194, 142}

\newcommand{\nar}{\textsc{NAR}\xspace}
\newcommand{\ar}{\textsc{AR}\xspace}
\newcommand{\mt}{\textsc{MT}\xspace}
\newcommand{\nmt}{\textsc{NMT}\xspace}
\newcommand{\imputer}{\textsc{Imputer}\xspace}
\newcommand{\ctc}{\textsc{CTC}\xspace}
\newcommand{\prism}{\textsc{PRISM}\xspace}

\newcommand{\bleu}{\textsc{BLEU}\xspace}

\newcommand{\frs}{\textsc{FRS}\xspace}
\newcommand{\bp}{\textsc{BP}\xspace}

\usepackage{xcolor}
\newcommand{\ensuretext}[1]{#1}
\newcommand{\samarker}{\ensuretext{\textcolor{orange}{\ensuremath{^{\textsc{S}}_{\textsc{A}}}}}}

\newcommand{\jkmarker}{\ensuretext{\textcolor{green}{\ensuremath{^{\textsc{J}}_{\textsc{K}}}}}}

\newcommand{\mycomment}[3]{}

\newcommand{\sa}[1]{\mycomment{\samarker}{#1}{orange}}

\newcommand{\jk}[1]{\mycomment{\jkmarker}{#1}{green}}


%% file: Introduction.tex
\section{Introduction}

Non-autoregressive (\nar) models generate output tokens in parallel instead of sequentially, achieving significantly faster inference speed that no longer depends on sequence length. They depend heavily on sequence-level knowledge distillation to reach the quality of \ar models \cite{gu2018nonautoregressive}. 
As the notion of \nar has expanded to include semi-\nar models that generate their outputs in multiple steps, each time generating several tokens non-autoregressively~\cite{lee-etal-2018-deterministic,ghazvininejad2019mask}, we have begun to see cases where \nar matches the quality of \ar.
Most prior works have tested the performance of \nar models on a handful of benchmarks of selected language pairs like German (De), Chinese (Zh), Romanian (Ro).
To efficiently expand this set of languages, it makes sense to begin exploring multilingual \nar translation models.

Multilingual machine translation models \cite{dong-etal-2015-multi, johnson-etal-2017-googles} translate between multiple languages. They have better parameter-efficiency than building one bilingual model per language pair, and they are able to transfer knowledge from high-resource languages to low-resource ones. They have become an attractive solution for expanding the language coverage of \ar models \cite{aharoni2019massively, fan2021beyond}. The capability of doing multilingual modeling is a major feature of the \ar regime, and it is one that we should seek to maintain in \nar models.


It is unclear to what extent the properties of multilingual \ar models apply to \nar models.
Do related languages help each other (positive transfer) as easily?
Do unrelated languages interfere with one another (negative transfer) to the same extent?
Since \nar models tend to trade target-side modeling for improved modeling of the source, the answer to both questions is unclear.
Furthermore, \nar modeling raises a new issue of multilingual distillation.
To retain the training-time efficiency of multilingual modeling, it is crucial that \nar works well with multilingual teachers; otherwise, the prospect of training many bilingual teachers would greatly increase the effective training cost.
It may actually be the case that multilingual teachers are better suited than bilingual ones, as the effective capacity reduction may result in less complex~\cite{zhou2019understanding} and less multimodal outputs~\cite{gu2018nonautoregressive}.



We present an empirical study of multilingual \nar modeling.
Taking CTC~\cite{libovicky-helcl-2018-end} as our canonical \nar method, and Imputer~\cite{saharia2020non} as our canonical semi-\nar model, we study how they respond to multilinguality in a 6-language scenario designed to emphasize negative transfer, as well as two-language scenarios designed to emphasize positive transfer. In doing so, we make the following contributions:
\begin{enumerate}
    \item We show that multilingual \nar models suffer more from negative transfer and benefit less from postive transfer than \ar models.
    \item We fit a scaling law for our 6-language \nar scenario, showing that this trend continues as model size increases.
    \item We demonstrate that multilingual \nar performs equally well with multilingual and bilingual teachers, even in scenarios where the multilingual teacher has lower BLEU.
\end{enumerate}
Unfortunately, our results indicate that the time is not quite right for multilingual \nar, as least for the models studied here, but our analysis should help future efforts in this space.

%% file: Method.tex

\section{Non-Autoregressive Multilingual \nmt}

Let, $D^l = {(x, y) \in X \times Y}$ denote the bilingual corpus of a language pair, $l$. Given an input sequence $x$ of length $T'$, an \ar model \cite{bahdanau2015neural, vaswani2017attention} predicts the target $y$ with length $T$ sequentially based on the conditional distribution
$p(y_t\mid y_{<t}, x_{1:T'}; \theta)$. \nar models assume conditional independence in the output token space; that is, they model $p(y_{1:T}\mid x_{1:T'}; \phi)$. Due to this conditional independence assumption, training \nar models directly on the true target distribution leads to degraded performance \cite{gu2018nonautoregressive}. Hence, \nar models are typically trained with sequence-level knowledge distillation~\cite{kim-rush-2016-sequence} to reduce the modeling difficulty. 

\input{tables/training_data_stats}

\subsection{Non-Autoregressive \nmt with \ctc}

 In this work, we focus on \nar modelling via \ctc \cite{graves2006connectionist} due to its superior performance on \nar generation and the flexibility of variable length prediction \cite{libovicky-helcl-2018-end, saharia2020non, gu-kong-2021-fully}. 
 
\ctc models an alignment $a$ that provides a mapping between a sequence of predicted and target tokens. Alignments can be constructed by inserting special \textit{blank tokens} ("\_") and token repetitions into the target sequence. The alignment is monotonic with respect to the target sequence and is always the same length as the source sequence $x$. However, in \mt, the target sequence $y$ can be longer than the source sequence x. This is handled via upsampling the source sequence $x$, to $s$ times its original length.  An alignment is valid only if when collapsed, i.e., merging repeated tokens and removing blank tokens, it results in the original target sequence. 
 The \ctc loss marginalizes over all possible valid alignments $\Gamma(y)$ compatible with the target $y$ and is defined as:
 \begin{equation*}
     p(y\mid x) = \sum_{a\in \Gamma(y)} p(a_{1:T'}\mid x_{1:T'}; \phi). 
 \end{equation*}
Note that each alignment token $a_{t'}$ is modeled independently. This conditional independence allows \ctc to predict the single most likely alignment non-autoregressively at inference time, which can then be efficiently collapsed to an output sequence. This same independence assumption enables efficient minimization of the \ctc loss via dynamic programming \cite{graves2006connectionist}.  While \ctc enforces monotonicity between the alignment and the target, it does not require any cross- or self-attention layers inside the model to be monotonic. Hence, \ctc should still be able to model language pairs with different word orders between the source and the target sequence.
 Following \citet{saharia2020non}, we train encoder-only \ctc models, using a stack of self-attention layers to map the source sequence directly to the alignments.  


\subsection{Iterative Decoding with Imputer}

\imputer \cite{saharia2020non} extends \nar \ctc modeling by iterative refinement \cite{lee-etal-2018-deterministic}. At each inference step, it conditions on a previous partially generated alignment to emit a new alignment. While \imputer, like \ctc, generates all tokens at each inference step, only a subset of these tokens are selected to generate a partial alignment, similar to iterative masking approaches~\cite{ghazvininejad2019mask}. 
This is achieved by training with marginalization over partial alignments:
 \begin{equation*}
     p(y\mid x) = \sum_{a\in \Gamma(a)} p(a\mid  a_{\text{Mask}}, x; \phi), 
 \end{equation*}
where $a_{\text{Mask}}$ is a partially masked input-alignment. 
At training time, the $a_{\text{Mask}}$ alignment is generated using a \ctc model trained on the same dataset, and its masked positions are selected randomly.
This training procedure enables \imputer to iteratively refine a partial alignment over multiple decoding steps at inference time - consuming its own alignments as input to the next iteration.
With $k>1$ decoding steps, the \imputer becomes \emph{semi}-autoregressive, requiring $k$ times more inference passes than pure \ctc models. 

\imputer differs from Conditional Masked Language Modeling (\textsc{CMLM}) \cite{ghazvininejad2019mask} in that it utilizes the \ctc loss instead of the standard cross-entropy loss, removing the need for explicit output length prediction. Also, \imputer is an encoder-only model that makes one prediction per source token, just like \ctc. The cross-attention component from encoder-decoder is replaced by a simple sum between the embeddings of the source sequence and the input alignment ($a_{\text{Mask}}$) before the first self-attention layer.\footnote{We experimented with an encoder-decoder variant of \imputer but it did not change the overall output quality in multilingual scenarios or otherwise.}


\subsection{Multilingual Modeling}
Multilingual \mt \cite{dong-etal-2015-multi, johnson-etal-2017-googles} extends bilingual \mt by training a single model with datasets from multiple language pairs, ${\{D^l\}}_{l=1}^L$.   
\jk{explain briefly what's needed to make them multilingual, multi-task vs multi-domain, temperature sampling and joint vocabularies, maybe some insights from multi-GLAT to set expectations} 
To enable multilingual modelling in both \ar and \nar models, we prepend each source sequence with the desired target language tag ($<$2tgt$>$) and generate a shared vocabulary across all languages \cite{johnson-etal-2017-googles}. The model encodes this tag as any other vocabulary token, and can use this to guide the generation of the output sequence in the desired target language.

\subsection{Efficiency}
\paragraph{Inference} We refrain from wallclock inference time measurements since these are dependent on implementation, low-level optimization and machines \cite{dehghani2021efficiency}, and instead compare generation speed in terms of the number of tokens that get generated per iteration \cite{kreutzer-etal-2020-inference}, which is $<1$ for \ar models,\footnote{$1$ for greedy search, $<1$ to account for multiple hypotheses that have to be considered and expanded in every step for beam search.}
$N$ for fully non-autoregressive models like \ctc and $k$ for iterative semi-autoregressive models like \imputer. We acknowledge that other factors like model-depth play a role for inference time, but we assume that both \nar and \ar models can be optimized for this aspect \cite{kasai2020deep}. 
\paragraph{Training} At training time, NAR models are less efficient than \ar models because their quality depends on distillation \cite{gu-kong-2021-fully}. Extra cost is incurred to train a teacher model (usually \ar) and to use it to decode the training set.
\paragraph{Multilinguality} As discussed above, multilingual models have the advantage of multi-tasking over language pairs, so that a single multilingual model can replace several bilingual models. Thanks to transfer across languages, model size usually needs to be increased less than $m$-fold for modeling $m$ languages instead of a single one.

Considering all of the above factors, an ideal model requires only a few iterations (decoder passes or steps), 
requires no teacher, and covers several languages, while incurring the smallest drop in quality compared to less efficient models.
\ctc is desirable as it uses only one pass, while Imputer gives up some efficiency to improve quality.
Both require a teacher, but we can try to reduce teacher training costs through distillation.

%% file: tables/training_data_stats.tex
\begin{table*}[!t]
    \centering
    \scalebox{0.83}{
    \begin{tabular}{lcrccccc}
    \rowcolor{gray!10}
    \textbf{\textsc{}} & \textbf{\textsc{Tgt Word Order}}& \textbf{\textsc{Size}}& \textbf{\textsc{Script Difference}} & \textbf{\textsc{White Space}} & \textbf{\textsc{Src Length}}  & \textbf{\textsc{Tgt Length}} \\
    \addlinespace[0.2cm]
    
     \textbf{\textsc{EN-KK}} & SOV & $150$K & \cmark   & \cmark & $26.7$ & $20.0$\\
      \addlinespace[0.2cm] 
      
    \textbf{\textsc{EN-DE}} & SVO/SOV & $4.6$M  & \xmark & \cmark & $25.7$ & $24.3$ \\
      \textbf{\textsc{EN-PL}} &  SVO & $5$M & \xmark & \cmark & $16.2$ & $14.6$ \\
        \textbf{\textsc{EN-HI}} & SOV & $8.6$M & \cmark  & \cmark & $18.3$ & $19.8$\\
        
         \addlinespace[0.2cm]
        
         \textbf{\textsc{EN-JA}} & SOV & $17.9$M & \cmark  & \xmark& $21.4$  & $25.9$ \\
       \textbf{\textsc{EN-RU}} & Free & $33.5$M & \cmark  & \cmark & $23.2$ & $21.5$\\
     \textbf{\textsc{EN-FR}} & SVO  & $38.1$M & \xmark & \cmark &  $29.2$ &  $32.8$ \\
    \end{tabular}}
    \caption{Details on training data used. Target word orders are the ones that are dominating within the language according to \cite{wals}, but there may be sentence-specific variations. English follows predominantly SVO (Subject-Verb-Object) order. Size is measured as the number of parallel sentences in the training data. Source (Src) and Target (Tgt) length are averaged across sentences after word-based tokenization.  \jk{maybe add domain column} \jk{maybe summarize lengths columns in length ratio column}}
    \label{tab:training_data}
\end{table*}

%% file: ExperimentalSetup.tex
\input{tables/all_en_xx_results}

\section{Experimental Setup}
\jk{maybe an intro sentence what we care for in our setup}
\paragraph{Data} We perform our experiments on six language pairs, translating from English into WMT-14  German (de)\footnote{\url{http://www.statmt.org/wmt14/translation-task.html}}, WMT-15 French (fr)\footnote{\url{http://www.statmt.org/wmt15/translation-task.html}}, WMT-19 Russian (ru)\footnote{\url{http://www.statmt.org/wmt19/translation-task.html}}, WMT-20 Japanese (ja), WMT-20 Polish (pl)\footnote{\url{https://www.statmt.org/wmt20/translation-task.html}} and Samanantar Hindi (hi) \cite{ramesh2021samanantar}. 
We also use WMT-19 English-Kazakh (kk)\footnote{\url{https://www.statmt.org/wmt19/translation-task.html}} in Section~\ref{sec:transfer}.
The sizes and properties of the datasets are listed in Table~\ref{tab:training_data}. Target word order and the writing script differ across these language pairs. We consider translating from English as this is more interesting and difficult direction. 
\sa{Need to add kazakh is only used for transfer experiments}

We use SentencePiece \cite{kudo2018sentencepiece} to generate a shared subword vocabulary for the source and target language pairs. The proportion of sub-words allocated for each language depends on the size of the language in the combined training data. 

\paragraph{Evaluation Metrics} We evaluate translation quality via
\bleu \cite{papineni2002bleu} as calculated by Sacrebleu~\cite{post-2018-call}.
For En-Ja we measure Character-level \bleu to be independent of specific tokenizers.

\paragraph{Architecture} We train the \imputer model using the same setup as described in \citet{saharia2020non}: We follow their base model with $d_{model}$ = $512$, $d_{hidden}$ = $2048$, $n_{heads}$ = $8$, $n_{layers}$ = $12$, and $p_{dropout}$ = $0.1$. \ar models follow Transformer-base~\cite{vaswani2017attention} and have similar parameter counts. We train both models using Adam with learning rate of $0.0001$. We train \ctc models with a batch size of $2048$ and $8192$ sentences for $300$K steps for the bilingual and multilingual models respectively. We train the \imputer using \ctc loss using a  Bernoulli masking policy for next $300$K steps with a batch size of $1024$ and $2048$ sentences for the bilingual and multilingual models respectively. We upsample the source sequence by a factor of $2$ for all our experiments.\footnote{While increasing the upsampling ratio can provide a larger alignment space, we do not vary the upsampling ratio due to small difference in the performance of the resulting \nar models (See Table 6, \citet{gu-kong-2021-fully}). } We pick the best checkpoint based on validation \bleu for bilingual models and use the last checkpoint for multilingual models.

\paragraph{Distillation} We apply sequence-level knowledge distillation \cite{kim-rush-2016-sequence} from \ar teacher models as widely used in \nar\ generation \cite{gu2018nonautoregressive}. Specifically, when training the \nar models, we replace the reference sequences during training with translation outputs from Transformer-Big \ar teacher model with beam = 4. We also report the quality of the \ar teacher models. We experiment with two types of teachers, bilingual and multilingual. 

%% file: tables/all_en_xx_results.tex
\begin{table*}[t]
\centering
 \setlength\tabcolsep{3pt}
\scalebox{0.95}{
\begin{tabular}{llcllllll|r}
 \toprule
  \textbf{\textsc{Model}} &  \textbf{\textsc{Teacher}} & $N_{gen}$ &  \textsc{EN-FR}  &  \textsc{EN-DE} &  \textsc{EN-PL} &  \textsc{EN-RU} &  \textsc{EN-HI} &  \textsc{EN-JA}  & \textsc{Avg.}    \\
     \midrule
     \ar-big & & \multirow{1}{*}{ $<1$ }  &38.8 &	29.0 &	21.4 &	27.2 &	34.6 &	35.4 & 31.1\\
       multi-\ar-big & & &38.5 &	27.0  &	21.6 &	25.3 &	32.6 &	33.6 & 29.3\\
        \addlinespace[0.2cm]
        \rowcolor{gray!10}
    \textbf{\textit{Bilingual Models}} \\
      \addlinespace[0.2cm]
     \ar-base & &  $<1$  & 38.2 &	27.6 &	21.2 &	26.2 &	33.8 &	34.8 & 30.3\\
      \addlinespace[0.2cm]
    \multirow{2}{*}{\ctc}   & \ar-big &   \multirow{2}{*}{N}  & 35.7	&25.2 &	18.0&	21.4&	31.6&	31.6 & 27.3\\
      & multi-\ar-big & & 35.1 &	24.0 &	17.7 & 20.8	 &	30.8 & 28.9 & 26.2 \\
      \addlinespace[0.2cm]
     \imputer & \ar-big &  8 &  38.5 &	27.2 &	21.2 &	25.6 &	32.0	 & 32.0 & 29.4\\
 \addlinespace[0.2cm]
 \rowcolor{gray!10}
 \textbf{\textit{Multilingual Models}}  \\
 \addlinespace[0.2cm]
multi-\ar-base & &  $<1$ & 35.2 &	24.8 &	19.7 &	23.2 &	30.8 &	31.2  & 27.5\\
      \addlinespace[0.2cm]
     \multirow{2}{*}{\ctc} & \ar-big &   \multirow{2}{*}{N} & 31.6 &	20.5 &	13.0 &	17.7 &	28.2 &	28.1 & 23.2\\
      &multi-\ar-big&  &31.2 &	20.5 &	13.7 &	18.0 &	27.8 &	27.5 & 23.1 \\
      \addlinespace[0.2cm]
          \multirow{2}{*}{\imputer} & \ar-big & \multirow{2}{*}{8}   & 34.4	& 22.8 &	14.9 &	21.3 &	29.9 & 29.6 & 25.5 \\
      & multi-\ar-big & & 34.1 & 21.2 & 16.4 & 21.7 & 29.9 & 27.9 & 25.2 \\

\bottomrule
 \end{tabular}}
\caption{Multilingual and Bilingual \ar and \nar models trained on $English \rightarrow X$ direction. }\label{tab:main_from_english} 
\end{table*}

%% file: Results.tex
\section{Negative Transfer Scenario}

Our main experiment compares English-to-X models for the six high-resource languages in Table~\ref{tab:training_data}. These languages are typologically diverse, and each have enough data so that we do not expect them to benefit substantially from positive transfer. We use this scenario to test the impact of multilingual teachers, and to measure each paradigm's ability to model several unrelated languages. Results are shown in Table~\ref{tab:main_from_english}.


\subsection{Multilingual Teacher Comparison} 
The top two rows of Table~\ref{tab:main_from_english} show that in this negative transfer scenario, multilingual teachers have substantially reduced \bleu compared to bilingual teachers.
However, as we look at the impact on bilingual students, we see that \ctc models trained from the multilingual teacher, {\ttfamily multi-\ar-big}, do not reflect the entirety of this drop in teacher quality when compared to training with the bilingual {\ttfamily \ar-big}. An average teacher gap of $-1.8$ \bleu is mapped to $-1.1$ in the corresponding students.
The comparison becomes more interesting as we shift to multilingual students:
multilingual CTC does not suffer at all from having a multilingual teacher (average \bleu gap of $-0.1$), and multilingual Imputer likewise suffers very little ($-0.3$).
These three results taken together suggest that datasets distilled from multilingual models are likely simpler and easier to model non-autoregressively, which makes up for their lower \bleu.
We explore this hypothesis further in Section~\ref{sec:analysis_distill}. 
We hope that highly multilingual models, trained with similar target language pairs to exhibit positive transfer \cite{tan2019multilingual}, might be yet better suited to serve as teachers for multilingual \nar models, which we leave to future work. 


\sa{Missing a stronger result: if we train multilingual \nar on distilled datasets with similar quality as bilingual, do we improve over bilingual models?}
\subsection{Multilingual Model Comparison} 

Returning to the ``Bilingual Models'' section of Table~\ref{tab:main_from_english} with {\ttfamily \ar-big} teachers, we can see that we have reproduced the expected results of \citet{saharia2020non}. 
Bilingual \ctc does well for a fully \nar method, but does not come close to \ar quality.
\imputer ably closes the gap with \ar, surpassing or coming within 0.2 \bleu of the \ar-base models on $3$ out of $6$ language pairs, with the largest gap in performance for the distant En-Ja. Does this story hold as we move to multilingual \nar students?


To understand each model's multilingual capabilities, we can compare its bilingual performance to its multilingual performance.
Comparing AR-base to multilingual AR-base gives us a baseline average drop of $-2.8$ \bleu , confirming that this is indeed a difficult multilingual scenario that leads to negative transfer.
Comparing bilingual \ctc to multilingual \ctc, both with AR-big teachers, we see an average drop of $-4.1$.
This larger drop indicates that \ctc suffers more from negative transfer than its \ar counterpart.
We hypothesize that \ctc needs more capacity compared to the \ar model to achieve similar multilingual performance, motivating our scaling law experiments in Section~\ref{sec:scaling}.
Performing the same bilingual-to-multilingual comparison for \imputer shows a similar $-3.9$ average drop due to negative transfer. So although \imputer is indeed substantially better than \ctc, it does not seem to be necessarily better suited for multilingual modeling in this difficult scenario.


%% file: Analysis.tex

\input{tables/data_differences}

\subsection{How do the distilled datasets differ?}
\label{sec:analysis_distill}

Table~\ref{tab:data_difference} summarizes different statistics for the original  ($R$) and distilled datasets from both multilingual ($M$) and bilingual ($B$) \ar teacher models. We report the number of types and average sequence length (in tokens) for the target side of the dataset. We compute the complexity of the dataset based on probabilities from a statistical word aligner~\citep{zhou2019understanding}. The \frs \cite{talbot2011lightweight} score represents the average fuzzy reordering score over all the sentence pairs for the respective language pair as measured in \citet{xu-etal-2021-distilled}, with higher values suggesting that the target is more monotonic with the source sequence.  We also report \bleu for the distilled datasets relative to the original training corpora.

The datasets distilled from the bilingual \ar models ($B$) are shorter, less complex, have reduced lexical diversity (in number of types) and are more monotonic compared to the original corpora ($R$), which is aligned with prior work \cite{zhou2019understanding, xu-etal-2021-distilled}. Interestingly, for En-Ja, we observe that the distilled datasets are less monotonic than the original corpora.

The multilingual distilled datasets ($M$) have further reduced types, are shorter and less complex than the distilled datasets from bilingual teachers. \sa{From Colin's comments:  Focus on complexity only here as other differences are small} The resulting distilled datasets from the multilingual teacher model specifically have increased monotonicity (\frs) for the more distant language pairs, Japanese and Hindi. As shown in \citet{xu-etal-2021-distilled}, the reduced lexical
diversity and reordering complexity both help \nar learn better alignment between source and target, improving the translation quality of the outputs.

\subsection{Error Analysis}


In this section, we present a qualitative analysis to provide
some insights on how \nar models differ in output quality across different language pairs when trained in isolation (bilingual) or with other language pairs (multilingual).

\begin{figure}[h]
    \centering
  \includegraphics[width=\linewidth]{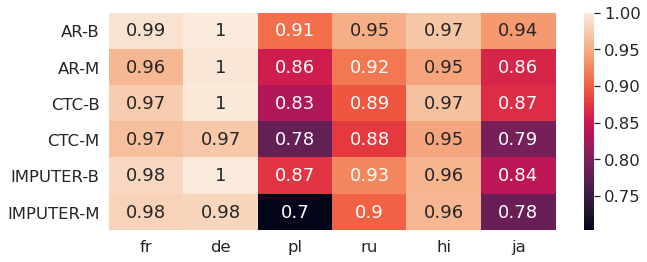}
    \caption{Brevity penalty (\bp) scores for all models for all the language pairs. ``-B'' and ``-M'' and bilingual and multilingual models respectively.}
    \label{fig:brevity}
\end{figure}

\paragraph{Effect of length} We show the brevity penalty scores from all the languages in Figure~\ref{fig:brevity}.   Among all the language pairs, both en-pl and en-ja have lowest brevity penalty scores. This could be attributed to the subject pronouns being dropped in both of these target languages. Multilingual modeling of most of the language pairs results in shorter outputs relative to bilingual models for both \ar and \nar models. 
While \imputer is generally able to improve the low brevity penalty values compared to \ctc models, they still lag behind \ar models, suggesting that length of the output might need to be controlled explicitly for these language pairs \cite{gu-kong-2021-fully}.


\paragraph{Invalid Words} 
Our manual inspection suggested that \ctc frequently generates \textit{invalid} words --- tokens that are not present in the target side of the training set or in the test set references. In the Hindi example below, the invalid (or made-up) word in the sentence is marked in red.
 
\begin{figure}[h]
    \centering
  \includegraphics[width=\linewidth]{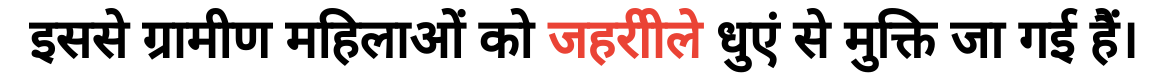}
    \label{fig:example}
\end{figure}
\vspace{-0.5cm}

\begin{figure}[h]
    \centering
  \includegraphics[width=\linewidth]{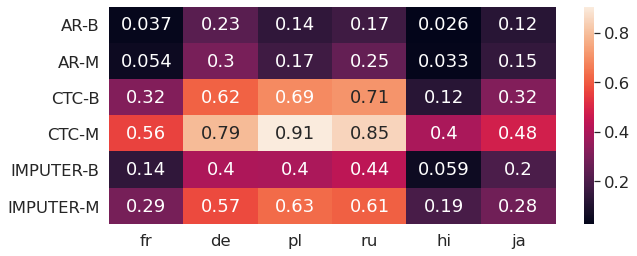}
    \caption{\% Invalid words observed in the outputs from all languages.``-B'' and ``-M'' and bilingual and multilingual models respectively.}
    \label{fig:invalid_words}
\end{figure}

 We compute the percentage of sequences that include at least one invalid word and report the statistics in Figure~\ref{fig:invalid_words}. \ctc generates many invalid words compared to both \ar (Average: 0.09-0.14) and \imputer (Average: 0.14-0.37), with multilingual modeling leading to an average increase in invalid words by $37\%$. We attribute this to the limited vocabulary of the model resulting in longer subword segmentation and the conditional independence assumption leading to unrelated adjacent subwords, which merge to create invalid words.

%% file: tables/data_differences.tex
\begin{table}[!htb]
\centering
\scalebox{0.95}{
\begin{tabular}{llrrr}
 \toprule
 \textsc{Property} & & R & B & M    \\
 \midrule
    \addlinespace[0.2cm]
   & & \textbf{\textsc{EN-FR}} & & \\
   \addlinespace[0.2cm]
  \textsc{\# Types}  & & $522$K& $430$K & $396$K\\
  \addlinespace[0.1cm]
   \textsc{Avg. Length}  && 32.8& 31.2& 29.2\\
   \addlinespace[0.1cm]
   \textsc{Complexity}  && 1.529& 1.167 & 0.944  \\
  \textsc{FRS}  && 0.463& 0.541 & 0.536  \\
   \addlinespace[0.1cm]
  \bleu (Train)  && - & 40.8 & 37.8    \\
  
     \addlinespace[0.2cm]
   & & \textbf{\textsc{EN-DE}} & & \\
   \addlinespace[0.2cm]
  \textsc{\# Types}  && $812$K& $616$K  & $573$K \\
  \addlinespace[0.1cm]
   \textsc{Avg. Length}  && 24.3 & 23.4 & 22.2 \\
   \addlinespace[0.1cm]
   \textsc{Complexity}  && 1.243 & 0.819& 0.709\\
  \textsc{FRS}  & & 0.490& 0.606 & 0.605  \\
   \addlinespace[0.1cm]
  \bleu (Train)  && - &  35.0 & 26.4     \\
  
     \addlinespace[0.2cm]
   & & \textbf{\textsc{EN-PL}} & & \\
   \addlinespace[0.2cm]
  \textsc{\# Types}  & & $636$K&  $516$K &$503$K\\
  \addlinespace[0.1cm]
   \textsc{Avg. Length}  && 14.6& 13.4 & 12.7  \\
   \addlinespace[0.1cm]
   \textsc{Complexity}  && 1.435&  0.942 & 0.591   \\
  \textsc{FRS}  & & 0.590& 0.678 & 0.695  \\
   \addlinespace[0.1cm]
  \bleu (Train)  && -&  26.3 & 22.0     \\
  
    \addlinespace[0.2cm]
   & & \textbf{\textsc{EN-RU}} & & \\
   \addlinespace[0.2cm]
  \textsc{\# Types}  && $636$K& $516$K &$503$K  \\
  \addlinespace[0.1cm]
   \textsc{Avg. Length}  && 21.5 &  20.5& 19.5 \\
   \addlinespace[0.1cm]
   \textsc{Complexity}  && 1.083& 0.882 & 0.819   \\
  \textsc{FRS}  && 0.640 & 0.719 & 0.716 \\
   \addlinespace[0.1cm]
  \bleu (Train)  && -& 43.2 & 40.0     \\
  
   \addlinespace[0.2cm]
   & & \textbf{\textsc{EN-HI}} & & \\
   \addlinespace[0.2cm]
  \textsc{\# Types}  && $346$K& $200$K & $185$K \\
  \addlinespace[0.1cm]
   \textsc{Avg. Length}  & & 19.8 & 18.8 & 17.8\\
   \addlinespace[0.1cm]
   \textsc{Complexity}  & & 1.438&  1.256 &1.138  \\
  \textsc{FRS}  && 0.347 &  0.363 & 0.366 \\
   \addlinespace[0.1cm]
  \bleu (Train)  && -&  34.6 & 28.0    \\
  
   \addlinespace[0.2cm]
   & & \textbf{\textsc{EN-JA}} & & \\
   \addlinespace[0.2cm]
  \textsc{\# Types}  && $547$K& $440$K & $402$K \\
  \addlinespace[0.2cm]
   \textsc{Avg. Length}  && 25.9&  23.5 & 22.2  \\
   \addlinespace[0.1cm]
   \textsc{Complexity}  & & 1.541 & 1.369 & 1.338\\
  \textsc{FRS}  && 0.344& 0.337 & 0.340  \\
   \addlinespace[0.1cm]
  \bleu (Train)  && -&  35.9 & 30.6     \\
\bottomrule
 \end{tabular}}
\caption{Comparison of datasets distilled from Bilingual (B) or Multilingual (M) \ar models on a subset of $1$M samples: Multilingual distilled datasets have fewer types, are less complex and more monotonic than bilingual distilled datasets, despite having lower \bleu. }\label{tab:data_difference} 
\end{table}

%% file: Transfer.tex
\section{Positive Transfer Scenario}
\label{sec:transfer}

In this section we present two experimental setups designed to study positive transfer for similar language pairs, where one or both languages have less than the ideal amount of data. 

\paragraph{English$\rightarrow$\{German, French\}}
To better isolate the effect of transfer via multilingual modelling, we simulate a resource-limited scenario by subsampling $1$M samples each from German and French and train bilingual and multilingual models for both \ar and \nar paradigms.
By pairing two related languages and subsampling down to smaller dataset sizes, we relax the capacity bottleneck and competition amongst the languages for parameters.

Table~\ref{tab:preliminary_setup} shows that \nar models benefit from training with multilingual language pairs in the resource-constrained scenario  --- all models exhibit positive transfer.
\imputer  achieves higher positive transfer than \ctc across both the language pairs, but lags behind the \ar multilingual model in en-fr. Note however that, for en-fr, the  bilingual \imputer is already ahead of the bilingual \ar model by $0.4$ \bleu. 

\input{tables/preliminary}

\paragraph{English$\rightarrow$\{Russian, Kazakh\}} We test the performance of the multilingual \nar model on the low-resource scenario of En$\rightarrow$Kk, where there is not sufficient clean training data to train an \ar model in the first place. We instead distill datasets from the publicly available multilingual autoregressive model, \prism~\cite{thompson-post-2020-paraphrase}.
We then pair with the high-resource Russian to encourage positive transfer to Kazakh.
Given the huge difference in dataset sizes for Russian and Kazakh (see Table~\ref{tab:training_data}), we sample training data from the two languages based on the dataset size scaled by a temperature value (T), $p_{l}^{1/T}$ \cite{arivazhagan2019massively}, where, $p_{l} = \frac{D_{l}}{\sum_k D_{k}}$. We experiment with multiple temperature values: 1, 3, 5, 10, 20, where $T=1$ implies $p_{ru}^{1/T}=0.995$, $p_{kk}^{1/T}=0.005$ and $T=20$ results in approximately, $p_{ru}^{1/T}=0.56$, $p_{kk}^{1/T}=0.45$. The best performance on validation set was using $T=5$ $(p_{ru}^{1/T}=0.75$, $p_{kk}^{1/T}=0.25)$.

As can be seen in Table~\ref{tab:main_kazakh}, both \ar and \ctc benefit from positive transfer when translating into Kazakh when trained with Russian. The \ctc-M model is able to improve (\bleu: +1.6) over the \ctc-B model but the overall quality of the outputs is very low compared to the teacher model (\bleu: -5.3).
It highlights that current \nar models do not perform well on very low-resource language pairs and might benefit from additional data augmentation strategies in addition to transfer from other similar language pairs.\footnote{ We do not train \imputer for Kazakh as the quality of both the distilled datasets and generated alignments from \ctc is very low.}
\input{tables/kazakh_results}

%% file: tables/preliminary.tex
\begin{table}[h]
\centering
 \setlength\tabcolsep{2pt}
\begin{tabular}{lrlrl}
 \toprule
  \textbf{\textsc{Model}} & \multicolumn{2}{c}{\textsc{EN-DE}} & \multicolumn{2}{c}{\textsc{EN-FR}} \\
  \midrule
    \addlinespace[0.2cm]
 \rowcolor{gray!10}
 \textbf{\textit{Bilingual Models}}  \\
 \addlinespace[0.2cm]
   \ar & \textbf{22.8} &  &27.7\\
   \ctc & 21.5  && 26.5\\
   \imputer & \textbf{22.8}  && \textbf{28.1} \\
   
    \addlinespace[0.2cm]
 \rowcolor{gray!10}
 \textbf{\textit{Multilingual Models}}  \\
 
   \ar & \textbf{24.3} & {\color{darkgreen}{$\mathbf{+1.5}$}}& \textbf{29.0} & {\color{darkgreen}{$\mathbf{+1.3}$}}\\
   \ctc & 22.1 & {\color{darkgreen}{$\mathbf{+0.6}$}}& 26.9 & {\color{darkgreen}{$\mathbf{+0.4}$}}\\
   \imputer & 23.7 & {\color{darkgreen}{$\mathbf{+1.3}$}} & 28.5 & {\color{darkgreen}{$\mathbf{+0.4}$}} \\

 \bottomrule
 \end{tabular}
\caption{Results on subsampled ($1$M) training dataset for German and French.}\label{tab:preliminary_setup} 
\end{table}

%% file: tables/kazakh_results.tex
\begin{table}[h]
\centering
 \setlength\tabcolsep{2pt}
\scalebox{0.90}{
\begin{tabular}{llrr}
 \toprule
  \textbf{\textsc{Model}}& \textbf{\textsc{Teacher}} &    \textsc{EN-KK}  & \textsc{EN-RU}   \\
  \midrule
  \prism & - & 8.9 &  27.0 \\
    \addlinespace[0.2cm]
     \textbf{\textit{Bilingual Models}}  \\
    \ar & \multirow{2}{*}{\prism}  & 4.4 & - \\
    \ctc &  & 1.2 & - \\
     \textbf{\textit{Multilingual Models}}  \\
      \ar & \multirow{2}{*}{\prism}  & 7.1 & 26.0 \\
        \ctc   &  & 2.8 & 20.4  \\

\bottomrule
 \end{tabular}}
\caption{Results on $English \rightarrow Kazakh$.}\label{tab:main_kazakh} 
\end{table}

%% file: Scaling.tex
\section{Impact of Model Scale }
\label{sec:scaling}
\sa{From Colin's comments: maybe revisit the capacity experiment to motivate this better}
\sa{From Colin's comments:  Need to explain when we include \imputer and when we don't }

Prior work has studied scaling laws for \mt to understand the relationship between the output quality (\bleu), the cross-entropy loss and the number of parameters used for training the model \cite{ghorbani2021scaling, gordon2021data}. Based on our hypothesis that \ctc might require more capacity than \ar models, we study the relationship between the number of parameters used for training the models and the development \bleu averaged across all the language pairs. 

    
We derive the relationship between \bleu and the number of parameters ($N$) directly from the scaling laws proposed in \citet{gordon2021data} and \citet{ghorbani2021scaling} as follows:
    \sa{From Colin's comments:need to define variables and flesh out more}
    \begin{equation*}
    \begin{split}
    L(N) \approx L_0 + \alpha_n (1/N)^{\alpha_k} \textrm{ \citep{ghorbani2021scaling}} \\
    \bleu(L) \approx C {\rm e}^{-kL} \textrm{ \citep{gordon2021data}}  \\
    \bleu(N) \approx a {\rm e}^{-b(1/N)^c} \textrm{ (this work)}\\
    \end{split}
\end{equation*}
where $L$ is the test loss, $\{\alpha_n, \alpha_k, L_0, C, k\}$ are fitted parameters from previous power laws, and $\{a, b, c$\} are the collapsed fitted parameters of our power law.  \citet{ghorbani2021scaling}'s $L_0$ corresponds to the irreducible loss of the data, which becomes $a$ in our formulation.


\paragraph{Setup}  We train seven different models with varying capacity using uniform scaling for both \ar and \ctc models (see below). We use the same number of layers and model dimension and train both \ar and \ctc models on distilled datasets from bilingual teacher,  {\ttfamily \ar-big}, to make a fair comparison.\footnote{We acknowledge that this setup might still be unfair to both \ctc and \ar models as the former has architectural differences (e.g. lack of cross-attention and distinction between encoding and decoding layers), whereas \ar models might be limited by the quality of their teacher models.}

\input{tables/models}

\begin{figure}[htb!]
\begin{subfigure}{0.5\textwidth}
  \centering
  \includegraphics[width=0.95\linewidth]{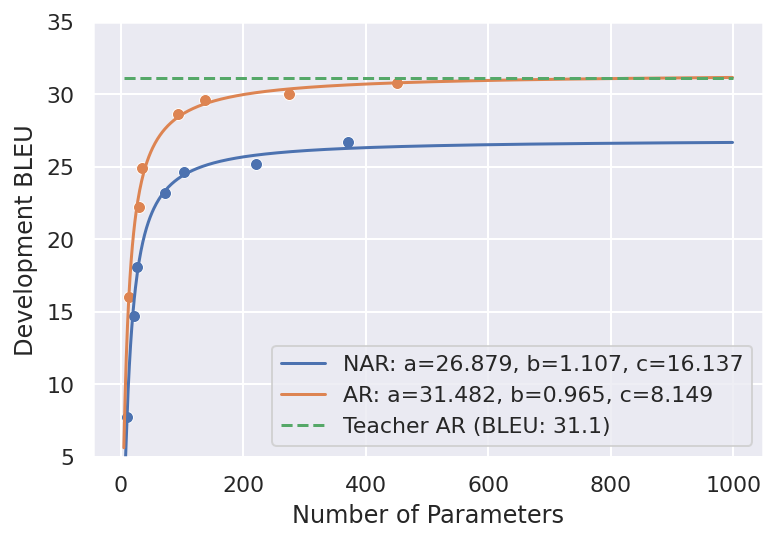}
  \caption{Scaling law fit ($R^2$ AR: 0.99, $R^2$ AR: 0.99).}
\end{subfigure}%
 \caption{ Development BLEU versus number of parameters in millions, with fitted power-law curves. \ctc requires relatively more parameters to achieve the same performance as \ar models. } \label{fig:curve_fit}
\end{figure}

\paragraph{Results} 
Figure~\ref{fig:curve_fit} shows the fitted parameters using the model derived using scaling laws respectively --- the scaling-law based model is almost able to perfectly describe the relationship between the number of parameters and the development \bleu ($R^2$ AR: 0.99). When the number of parameters are less than $10$M, both \ar and \ctc model result in approximately similar quality outputs. However, as the number of parameters increases, the gap in \bleu also increases, suggesting that with sufficient number of parameters \ar models are able to generate higher quality outputs due to the conditional dependence between tokens. We can also see that \ctc needs many more parameters to achieve comparable \bleu to \ar models and plateaus early at a \bleu of $26.7$, while \ar models plateau at $30.8$. By projecting the curves out to 1 billion parameters, we can show that increasing the capacity of \nar is insufficient to reach the quality of \ar models.

%% file: tables/models.tex
\begin{table}[h]
\centering
 \setlength\tabcolsep{5pt}
\begin{tabular}{clc}
\rowcolor{gray!10}
\# of Layers & &  Model Size  \\
 6 &&	128 \\
6&&	256 \\
12 &&	256 \\
12	&& 512 \\
24	&& 512 \\
12	&& 1024 \\
24	&& 1024 \\
 
 \end{tabular}
\caption{The feed-forward size is $4$ times the size of the model.  All \ar models have equal number of encoder and decoder layers. The number of attention heads is given by $(8  / (512 / Model Size)) $. }\label{tab:model_variants} 
\end{table}

%% file: RelatedWork.tex
\section{Related Work}


\paragraph{Non-Autoregressive \mt} Multiple approaches with varying architectures \cite{gu2018nonautoregressive,Gu2019Levenshtein, chan2020imputer, xu2021editor}, custom loss functions \cite{ghazvininejad2020aligned,du2021order} and training strategies \cite{ghazvininejad2019mask, qian-etal-2021-glancing} have been used to enable parallel generation of output tokens for MT. While most of the prior work focuses on better utilization of distilled datasets, we focus on evaluating and understanding the impact of using multilingual distilled datasets for \nar training. 

\paragraph{Multilingual \mt} There has been a lot of interest in the \ar literature on understanding and proposing models that enable translation between more than two language pairs \cite{dabre2020survey}. Both supervised and unsupervised \cite{sun-etal-2020-knowledge-distillation} learning in \mt have benefitted from training with multiple languages, especially those that have very litte \cite{siddhant-etal-2020-leveraging} to no training data \cite{zhang-etal-2020-improving}. However, multilingual modelling has not yet received any attention in the \nar literature. Concurrent to our work, \citet{anonymous2022nonautoregressive} investigate a non-autoregressive multilingual machine translation model with a code-switch decoder. They show that adding code-switched back-translation data to the training of multilingual models improves performance. Our work instead focuses on understanding multilinguality for both the student and the teacher model in the context of \nar training without using any additional data augmentation strategies.

\paragraph{Distillation} Sequence-level knowledge distillation is one of the key ingredient in the training of \nar models. Recent works have focused on understanding the success of knowledge distillation in \nar training. \citet{zhou2019understanding} show that distilled datasets have reduced complexity compared to original bitext which is suitable for  \nar training. \citet{xu-etal-2021-distilled} further show that different types of complexity, i.e. reducing lexical diversity and reordering degree have different impacts on the training. \citet{voita-etal-2021-language} argue that the complexity of the dataset increases as the training of the \ar model progresses and use this to improve the performance of \nar model by distilling from an earlier checkpoint.  In this work, we focus on understanding the impact on quality and complexity of distilled datasets from multilingual and bilingual \ar teacher models.

\paragraph{Scaling Laws} While large scale models improve performance, it is practically impossible, time consuming and expensive to train the different variants of the model given different architectures and dataset sizes based on the amount of compute available. Recent works have derived empirical scaling laws that govern the relationship between the performance of the model and these factors \cite{Kaplan2020ScalingLF, hernandez2021scaling,bahri2021explaining,  gordon2021data, ghorbani2021scaling}. However, these scaling laws have not yet been studied for multilingual \mt which we explore in our work. 

%% file: main.bbl
\begin{thebibliography}{45}
\expandafter\ifx\csname natexlab\endcsname\relax\def\natexlab#1{#1}\fi

\bibitem[{Aharoni et~al.(2019)Aharoni, Johnson, and
  Firat}]{aharoni2019massively}
Roee Aharoni, Melvin Johnson, and Orhan Firat. 2019.
\newblock Massively multilingual neural machine translation.
\newblock In \emph{Proceedings of NAACL-HLT}, pages 3874--3884.

\bibitem[{Anonymous(2022)}]{anonymous2022nonautoregressive}
Anonymous. 2022.
\newblock \href {https://openreview.net/forum?id=5HvpvYd68b}
  {Non-autoregressive models are better multilingual translators}.
\newblock In \emph{Submitted to The Tenth International Conference on Learning
  Representations}.
\newblock Under review.

\bibitem[{Arivazhagan et~al.(2019)Arivazhagan, Bapna, Firat, Lepikhin, Johnson,
  Krikun, Chen, Cao, Foster, Cherry et~al.}]{arivazhagan2019massively}
Naveen Arivazhagan, Ankur Bapna, Orhan Firat, Dmitry Lepikhin, Melvin Johnson,
  Maxim Krikun, Mia~Xu Chen, Yuan Cao, George Foster, Colin Cherry, et~al.
  2019.
\newblock Massively multilingual neural machine translation in the wild:
  Findings and challenges.
\newblock \emph{arXiv preprint arXiv:1907.05019}.

\bibitem[{Bahdanau et~al.(2015)Bahdanau, Cho, and Bengio}]{bahdanau2015neural}
Dzmitry Bahdanau, Kyung~Hyun Cho, and Yoshua Bengio. 2015.
\newblock Neural machine translation by jointly learning to align and
  translate.
\newblock In \emph{3rd International Conference on Learning Representations,
  ICLR 2015}.

\bibitem[{Bahri et~al.(2021)Bahri, Dyer, Kaplan, Lee, and
  Sharma}]{bahri2021explaining}
Yasaman Bahri, Ethan Dyer, Jared Kaplan, Jaehoon Lee, and Utkarsh Sharma. 2021.
\newblock Explaining neural scaling laws.
\newblock \emph{arXiv preprint arXiv:2102.06701}.

\bibitem[{Chan et~al.(2020)Chan, Saharia, Hinton, Norouzi, and
  Jaitly}]{chan2020imputer}
William Chan, Chitwan Saharia, Geoffrey Hinton, Mohammad Norouzi, and Navdeep
  Jaitly. 2020.
\newblock Imputer: Sequence modelling via imputation and dynamic programming.
\newblock In \emph{International Conference on Machine Learning}, pages
  1403--1413. PMLR.

\bibitem[{Dabre et~al.(2020)Dabre, Chu, and Kunchukuttan}]{dabre2020survey}
Raj Dabre, Chenhui Chu, and Anoop Kunchukuttan. 2020.
\newblock A survey of multilingual neural machine translation.
\newblock \emph{ACM Computing Surveys (CSUR)}, 53(5):1--38.

\bibitem[{Dehghani et~al.(2021)Dehghani, Arnab, Beyer, Vaswani, and
  Tay}]{dehghani2021efficiency}
Mostafa Dehghani, Anurag Arnab, Lucas Beyer, Ashish Vaswani, and Yi~Tay. 2021.
\newblock The efficiency misnomer.
\newblock \emph{arXiv preprint arXiv:2110.12894}.

\bibitem[{Dong et~al.(2015)Dong, Wu, He, Yu, and Wang}]{dong-etal-2015-multi}
Daxiang Dong, Hua Wu, Wei He, Dianhai Yu, and Haifeng Wang. 2015.
\newblock \href {https://doi.org/10.3115/v1/P15-1166} {Multi-task learning for
  multiple language translation}.
\newblock In \emph{Proceedings of the 53rd Annual Meeting of the Association
  for Computational Linguistics and the 7th International Joint Conference on
  Natural Language Processing (Volume 1: Long Papers)}, pages 1723--1732,
  Beijing, China. Association for Computational Linguistics.

\bibitem[{Dryer and Haspelmath(2013)}]{wals}
Matthew~S. Dryer and Martin Haspelmath, editors. 2013.
\newblock \href {https://wals.info/} {\emph{WALS Online}}.
\newblock Max Planck Institute for Evolutionary Anthropology, Leipzig.

\bibitem[{Du et~al.(2021)Du, Tu, and Jiang}]{du2021order}
Cunxiao Du, Zhaopeng Tu, and Jing Jiang. 2021.
\newblock Order-agnostic cross entropy for non-autoregressive machine
  translation.
\newblock \emph{arXiv preprint arXiv:2106.05093}.

\bibitem[{Fan et~al.(2021)Fan, Bhosale, Schwenk, Ma, El-Kishky, Goyal, Baines,
  Celebi, Wenzek, Chaudhary et~al.}]{fan2021beyond}
Angela Fan, Shruti Bhosale, Holger Schwenk, Zhiyi Ma, Ahmed El-Kishky,
  Siddharth Goyal, Mandeep Baines, Onur Celebi, Guillaume Wenzek, Vishrav
  Chaudhary, et~al. 2021.
\newblock Beyond english-centric multilingual machine translation.
\newblock \emph{Journal of Machine Learning Research}, 22(107):1--48.

\bibitem[{Ghazvininejad et~al.(2020)Ghazvininejad, Karpukhin, Zettlemoyer, and
  Levy}]{ghazvininejad2020aligned}
Marjan Ghazvininejad, Vladimir Karpukhin, Luke Zettlemoyer, and Omer Levy.
  2020.
\newblock \href {https://arxiv.org/abs/2004.01655} {Aligned cross entropy for
  non-autoregressive machine translation}.
\newblock \emph{CoRR}, abs/2004.01655.

\bibitem[{Ghazvininejad et~al.(2019)Ghazvininejad, Levy, Liu, and
  Zettlemoyer}]{ghazvininejad2019mask}
Marjan Ghazvininejad, Omer Levy, Yinhan Liu, and Luke Zettlemoyer. 2019.
\newblock Mask-predict: Parallel decoding of conditional masked language
  models.
\newblock In \emph{Proceedings of the 2019 Conference on Empirical Methods in
  Natural Language Processing and the 9th International Joint Conference on
  Natural Language Processing (EMNLP-IJCNLP)}, pages 6112--6121.

\bibitem[{Ghorbani et~al.(2021)Ghorbani, Firat, Freitag, Bapna, Krikun, Garcia,
  Chelba, and Cherry}]{ghorbani2021scaling}
Behrooz Ghorbani, Orhan Firat, Markus Freitag, Ankur Bapna, Maxim Krikun,
  Xavier Garcia, Ciprian Chelba, and Colin Cherry. 2021.
\newblock Scaling laws for neural machine translation.
\newblock \emph{arXiv preprint arXiv:2109.07740}.

\bibitem[{Gordon et~al.(2021)Gordon, Duh, and Kaplan}]{gordon2021data}
Mitchell~A Gordon, Kevin Duh, and Jared Kaplan. 2021.
\newblock Data and parameter scaling laws for neural machine translation.
\newblock In \emph{Proceedings of the 2021 Conference on Empirical Methods in
  Natural Language Processing}, pages 5915--5922.

\bibitem[{Graves et~al.(2006)Graves, Fern{\'a}ndez, Gomez, and
  Schmidhuber}]{graves2006connectionist}
Alex Graves, Santiago Fern{\'a}ndez, Faustino Gomez, and J{\"u}rgen
  Schmidhuber. 2006.
\newblock Connectionist temporal classification: labelling unsegmented sequence
  data with recurrent neural networks.
\newblock In \emph{Proceedings of the 23rd international conference on Machine
  learning}, pages 369--376.

\bibitem[{Gu et~al.(2018)Gu, Bradbury, Xiong, Li, and
  Socher}]{gu2018nonautoregressive}
Jiatao Gu, James Bradbury, Caiming Xiong, Victor~O.K. Li, and Richard Socher.
  2018.
\newblock \href {https://openreview.net/forum?id=B1l8BtlCb} {Non-autoregressive
  neural machine translation}.
\newblock In \emph{International Conference on Learning Representations}.

\bibitem[{Gu and Kong(2021)}]{gu-kong-2021-fully}
Jiatao Gu and Xiang Kong. 2021.
\newblock \href {https://doi.org/10.18653/v1/2021.findings-acl.11} {Fully
  non-autoregressive neural machine translation: Tricks of the trade}.
\newblock In \emph{Findings of the Association for Computational Linguistics:
  ACL-IJCNLP 2021}, pages 120--133, Online. Association for Computational
  Linguistics.

\bibitem[{Gu et~al.(2019)Gu, Wang, and Zhao}]{Gu2019Levenshtein}
Jiatao Gu, Changhan Wang, and Junbo Zhao. 2019.
\newblock \href {http://papers.nips.cc/paper/9297-levenshtein-transformer.pdf}
  {Levenshtein transformer}.
\newblock In H.~Wallach, H.~Larochelle, A.~Beygelzimer, F.~d\textquotesingle
  Alch\'{e}-Buc, E.~Fox, and R.~Garnett, editors, \emph{Advances in Neural
  Information Processing Systems 32}, pages 11179--11189. Curran Associates,
  Inc.

\bibitem[{Hernandez et~al.(2021)Hernandez, Kaplan, Henighan, and
  McCandlish}]{hernandez2021scaling}
Danny Hernandez, Jared Kaplan, Tom Henighan, and Sam McCandlish. 2021.
\newblock Scaling laws for transfer.
\newblock \emph{arXiv preprint arXiv:2102.01293}.

\bibitem[{Johnson et~al.(2017)Johnson, Schuster, Le, Krikun, Wu, Chen, Thorat,
  Vi{\'e}gas, Wattenberg, Corrado, Hughes, and
  Dean}]{johnson-etal-2017-googles}
Melvin Johnson, Mike Schuster, Quoc~V. Le, Maxim Krikun, Yonghui Wu, Zhifeng
  Chen, Nikhil Thorat, Fernanda Vi{\'e}gas, Martin Wattenberg, Greg Corrado,
  Macduff Hughes, and Jeffrey Dean. 2017.
\newblock \href {https://doi.org/10.1162/tacl_a_00065} {{G}oogle{'}s
  multilingual neural machine translation system: Enabling zero-shot
  translation}.
\newblock \emph{Transactions of the Association for Computational Linguistics},
  5:339--351.

\bibitem[{Kaplan et~al.(2020)Kaplan, McCandlish, Henighan, Brown, Chess, Child,
  Gray, Radford, Wu, and Amodei}]{Kaplan2020ScalingLF}
Jared Kaplan, Sam McCandlish, T.~J. Henighan, Tom~B. Brown, Benjamin Chess,
  Rewon Child, Scott Gray, Alec Radford, Jeff Wu, and Dario Amodei. 2020.
\newblock Scaling laws for neural language models.
\newblock \emph{ArXiv}, abs/2001.08361.

\bibitem[{Kasai et~al.(2020)Kasai, Pappas, Peng, Cross, and
  Smith}]{kasai2020deep}
Jungo Kasai, Nikolaos Pappas, Hao Peng, James Cross, and Noah Smith. 2020.
\newblock Deep encoder, shallow decoder: Reevaluating non-autoregressive
  machine translation.
\newblock In \emph{International Conference on Learning Representations}.

\bibitem[{Kim and Rush(2016)}]{kim-rush-2016-sequence}
Yoon Kim and Alexander~M. Rush. 2016.
\newblock \href {https://doi.org/10.18653/v1/D16-1139} {Sequence-level
  knowledge distillation}.
\newblock In \emph{Proceedings of the 2016 Conference on Empirical Methods in
  Natural Language Processing}, pages 1317--1327, Austin, Texas. Association
  for Computational Linguistics.

\bibitem[{Kreutzer et~al.(2020)Kreutzer, Foster, and
  Cherry}]{kreutzer-etal-2020-inference}
Julia Kreutzer, George Foster, and Colin Cherry. 2020.
\newblock \href {https://doi.org/10.18653/v1/2020.emnlp-main.465} {Inference
  strategies for machine translation with conditional masking}.
\newblock In \emph{Proceedings of the 2020 Conference on Empirical Methods in
  Natural Language Processing (EMNLP)}, pages 5774--5782, Online. Association
  for Computational Linguistics.

\bibitem[{Kudo and Richardson(2018)}]{kudo2018sentencepiece}
Taku Kudo and John Richardson. 2018.
\newblock Sentencepiece: A simple and language independent subword tokenizer
  and detokenizer for neural text processing.
\newblock In \emph{EMNLP (Demonstration)}.

\bibitem[{Lee et~al.(2018)Lee, Mansimov, and Cho}]{lee-etal-2018-deterministic}
Jason Lee, Elman Mansimov, and Kyunghyun Cho. 2018.
\newblock \href {https://doi.org/10.18653/v1/D18-1149} {Deterministic
  non-autoregressive neural sequence modeling by iterative refinement}.
\newblock In \emph{Proceedings of the 2018 Conference on Empirical Methods in
  Natural Language Processing}, pages 1173--1182, Brussels, Belgium.
  Association for Computational Linguistics.

\bibitem[{Libovick{\'y} and Helcl(2018)}]{libovicky-helcl-2018-end}
Jind{\v{r}}ich Libovick{\'y} and Jind{\v{r}}ich Helcl. 2018.
\newblock \href {https://doi.org/10.18653/v1/D18-1336} {End-to-end
  non-autoregressive neural machine translation with connectionist temporal
  classification}.
\newblock In \emph{Proceedings of the 2018 Conference on Empirical Methods in
  Natural Language Processing}, pages 3016--3021, Brussels, Belgium.
  Association for Computational Linguistics.

\bibitem[{Papineni et~al.(2002)Papineni, Roukos, Ward, and
  Zhu}]{papineni2002bleu}
Kishore Papineni, Salim Roukos, Todd Ward, and Wei-Jing Zhu. 2002.
\newblock Bleu: a method for automatic evaluation of machine translation.
\newblock In \emph{Proceedings of the 40th annual meeting of the Association
  for Computational Linguistics}, pages 311--318.

\bibitem[{Post(2018)}]{post-2018-call}
Matt Post. 2018.
\newblock \href {https://doi.org/10.18653/v1/W18-6319} {A call for clarity in
  reporting {BLEU} scores}.
\newblock In \emph{Proceedings of the Third Conference on Machine Translation:
  Research Papers}, pages 186--191, Brussels, Belgium. Association for
  Computational Linguistics.

\bibitem[{Qian et~al.(2021)Qian, Zhou, Bao, Wang, Qiu, Zhang, Yu, and
  Li}]{qian-etal-2021-glancing}
Lihua Qian, Hao Zhou, Yu~Bao, Mingxuan Wang, Lin Qiu, Weinan Zhang, Yong Yu,
  and Lei Li. 2021.
\newblock \href {https://doi.org/10.18653/v1/2021.acl-long.155} {Glancing
  transformer for non-autoregressive neural machine translation}.
\newblock In \emph{Proceedings of the 59th Annual Meeting of the Association
  for Computational Linguistics and the 11th International Joint Conference on
  Natural Language Processing (Volume 1: Long Papers)}, pages 1993--2003,
  Online. Association for Computational Linguistics.

\bibitem[{Ramesh et~al.(2021)Ramesh, Doddapaneni, Bheemaraj, Jobanputra, AK,
  Sharma, Sahoo, Diddee, Kakwani, Kumar et~al.}]{ramesh2021samanantar}
Gowtham Ramesh, Sumanth Doddapaneni, Aravinth Bheemaraj, Mayank Jobanputra,
  Raghavan AK, Ajitesh Sharma, Sujit Sahoo, Harshita Diddee, Divyanshu Kakwani,
  Navneet Kumar, et~al. 2021.
\newblock Samanantar: The largest publicly available parallel corpora
  collection for 11 indic languages.
\newblock \emph{arXiv preprint arXiv:2104.05596}.

\bibitem[{Saharia et~al.(2020)Saharia, Chan, Saxena, and
  Norouzi}]{saharia2020non}
Chitwan Saharia, William Chan, Saurabh Saxena, and Mohammad Norouzi. 2020.
\newblock Non-autoregressive machine translation with latent alignments.
\newblock In \emph{Proceedings of the 2020 Conference on Empirical Methods in
  Natural Language Processing (EMNLP)}, pages 1098--1108.

\bibitem[{Siddhant et~al.(2020)Siddhant, Bapna, Cao, Firat, Chen, Kudugunta,
  Arivazhagan, and Wu}]{siddhant-etal-2020-leveraging}
Aditya Siddhant, Ankur Bapna, Yuan Cao, Orhan Firat, Mia Chen, Sneha Kudugunta,
  Naveen Arivazhagan, and Yonghui Wu. 2020.
\newblock \href {https://doi.org/10.18653/v1/2020.acl-main.252} {Leveraging
  monolingual data with self-supervision for multilingual neural machine
  translation}.
\newblock In \emph{Proceedings of the 58th Annual Meeting of the Association
  for Computational Linguistics}, pages 2827--2835, Online. Association for
  Computational Linguistics.

\bibitem[{Sun et~al.(2020)Sun, Wang, Chen, Utiyama, Sumita, and
  Zhao}]{sun-etal-2020-knowledge-distillation}
Haipeng Sun, Rui Wang, Kehai Chen, Masao Utiyama, Eiichiro Sumita, and Tiejun
  Zhao. 2020.
\newblock \href {https://doi.org/10.18653/v1/2020.acl-main.324} {Knowledge
  distillation for multilingual unsupervised neural machine translation}.
\newblock In \emph{Proceedings of the 58th Annual Meeting of the Association
  for Computational Linguistics}, pages 3525--3535, Online. Association for
  Computational Linguistics.

\bibitem[{Talbot et~al.(2011)Talbot, Kazawa, Ichikawa, Katz-Brown, Seno, and
  Och}]{talbot2011lightweight}
David Talbot, Hideto Kazawa, Hiroshi Ichikawa, Jason Katz-Brown, Masakazu Seno,
  and Franz~Josef Och. 2011.
\newblock A lightweight evaluation framework for machine translation
  reordering.
\newblock In \emph{Proceedings of the Sixth Workshop on Statistical Machine
  Translation}, pages 12--21.

\bibitem[{Tan et~al.(2019)Tan, Chen, He, Xia, Qin, and
  Liu}]{tan2019multilingual}
Xu~Tan, Jiale Chen, Di~He, Yingce Xia, Tao Qin, and Tie-Yan Liu. 2019.
\newblock Multilingual neural machine translation with language clustering.
\newblock In \emph{Proceedings of the 2019 Conference on Empirical Methods in
  Natural Language Processing and the 9th International Joint Conference on
  Natural Language Processing (EMNLP-IJCNLP)}, pages 963--973.

\bibitem[{Thompson and Post(2020)}]{thompson-post-2020-paraphrase}
Brian Thompson and Matt Post. 2020.
\newblock \href {https://www.aclweb.org/anthology/2020.wmt-1.67} {Paraphrase
  generation as zero-shot multilingual translation: Disentangling semantic
  similarity from lexical and syntactic diversity}.
\newblock In \emph{Proceedings of the Fifth Conference on Machine Translation},
  pages 561--570, Online. Association for Computational Linguistics.

\bibitem[{Vaswani et~al.(2017)Vaswani, Shazeer, Parmar, Uszkoreit, Jones,
  Gomez, Kaiser, and Polosukhin}]{vaswani2017attention}
Ashish Vaswani, Noam Shazeer, Niki Parmar, Jakob Uszkoreit, Llion Jones,
  Aidan~N Gomez, {\L}ukasz Kaiser, and Illia Polosukhin. 2017.
\newblock Attention is all you need.
\newblock In \emph{Advances in neural information processing systems}, pages
  5998--6008.

\bibitem[{Voita et~al.(2021)Voita, Sennrich, and
  Titov}]{voita-etal-2021-language}
Elena Voita, Rico Sennrich, and Ivan Titov. 2021.
\newblock \href {https://aclanthology.org/2021.emnlp-main.667} {Language
  modeling, lexical translation, reordering: The training process of {NMT}
  through the lens of classical {SMT}}.
\newblock In \emph{Proceedings of the 2021 Conference on Empirical Methods in
  Natural Language Processing}, pages 8478--8491, Online and Punta Cana,
  Dominican Republic. Association for Computational Linguistics.

\bibitem[{Xu and Carpuat(2021)}]{xu2021editor}
Weijia Xu and Marine Carpuat. 2021.
\newblock Editor: An edit-based transformer with repositioning for neural
  machine translation with soft lexical constraints.
\newblock \emph{Transactions of the Association for Computational Linguistics},
  9:311--328.

\bibitem[{Xu et~al.(2021)Xu, Ma, Zhang, and Carpuat}]{xu-etal-2021-distilled}
Weijia Xu, Shuming Ma, Dongdong Zhang, and Marine Carpuat. 2021.
\newblock \href {https://doi.org/10.18653/v1/2021.findings-acl.385} {How does
  distilled data complexity impact the quality and confidence of
  non-autoregressive machine translation?}
\newblock In \emph{Findings of the Association for Computational Linguistics:
  ACL-IJCNLP 2021}, pages 4392--4400, Online. Association for Computational
  Linguistics.

\bibitem[{Zhang et~al.(2020)Zhang, Williams, Titov, and
  Sennrich}]{zhang-etal-2020-improving}
Biao Zhang, Philip Williams, Ivan Titov, and Rico Sennrich. 2020.
\newblock \href {https://doi.org/10.18653/v1/2020.acl-main.148} {Improving
  massively multilingual neural machine translation and zero-shot translation}.
\newblock In \emph{Proceedings of the 58th Annual Meeting of the Association
  for Computational Linguistics}, pages 1628--1639, Online. Association for
  Computational Linguistics.

\bibitem[{Zhou et~al.(2019)Zhou, Gu, and Neubig}]{zhou2019understanding}
Chunting Zhou, Jiatao Gu, and Graham Neubig. 2019.
\newblock Understanding knowledge distillation in non-autoregressive machine
  translation.
\newblock In \emph{International Conference on Learning Representations}.

\end{thebibliography}
